# Bridge Networks

## Relating Inputs through Vector-Symbolic Manipulations

Wilkie Olin-Ammentorp
Department of Medicine, Institute for Neural Computation,
University of California, San Diego
San Diego, CA, USA
wolinammentorp@ucsd.edu

Maxim Bazhenov
Department of Medicine, Institute for Neural Computation,
University of California, San Diego
San Diego, CA, USA
mbazhenov@ucsd.edu

## ABSTRACT

Despite rapid progress, current deep learning methods face a number of critical challenges. These include high energy consumption, catastrophic forgetting, dependance on global losses, and an inability to reason symbolically. By combining concepts from information theory and vector-symbolic architectures, we propose and implement a novel information processing architecture, the 'Bridge network.' We show this architecture provides unique advantages which can address the problem of global losses and catastrophic forgetting. Furthermore, we argue that it provides a further basis for increasing energy efficiency of execution and the ability to reason symbolically.

## 1 Introduction

Advances in artificial intelligence (AI) have provided a variety of methods to create powerful function approximators, which can be applied to tasks such as image classification and object detection [10, 24]. However, applying these systems to real-world applications such as self-driving cars remains a challenge as deep networks remain inefficient in terms of both data and energy usage when compared to their biological counterparts [14, 16, 21]. These challenges have led to the exploration of new avenues of computing for artificial intelligence, including neuromorphic systems and a renewed interest in symbolic processing or 'neurosymbolic' architectures [5, 9, 20].

In general, neuromorphic computing seeks to provide methods for efficient computing in a manner which more closely mimics biology than current methods. This can be accomplished by several means, including colocation of computing and memory elements and communication through sparse signals such as binary 'spikes.'

The current challenges in scaling AI beyond current hardware limits have led to an expansion in a number of neuromorphic hardware systems which utilize different principles to accelerate processing [1, 2, 6, 11, 17].

Neurosymbolic AI approaches seek to provide architectural solutions to challenges for deep networks such as data inefficiency and lack of transparency by augmenting connectionist solutions with symbolic reasoning abilities [9]. One goal of this approach is to allow systems to perform 'less than one shot learning,' where new classes can be learned with no new training examples. For instance, having learned to recognize horses and rhinoceroses, a neurosymbolic system should be able to recognize a picture of a unicorn given only the knowledge that this new class combines the features of other classes (a horse's body and rhino's horn) [22].

Vector-symbolic architectures provide a powerful basis for distributed and neurosymbolic computing. In general, these architectures utilize vectors in a high-dimensional space which can be related to one another through 'bundling' and 'binding' operations. 'Bundling' can be thought of as combining multiple input symbols into a single output symbol. This output remains measurably similar to its inputs and dissimilar to other data points; this is achieved through the high-dimensional space in which the vectors lie [15]. In this work, we utilize the Fourier Holographic Reduced Representation (FHRR) architecture as it is computationally efficient and can be executed efficiently via spiking architectures [8, 18, 19].

Towards the goal of developing an architecture which can utilize both neuromorphic and neurosymbolic principles, in this work we demonstrate a 'Bridge Network,' a neurally-inspired architecture which learns to compute via vector-symbolic representations.

## 2 Results

A Bridge network consists of multiple processing blocks which have access to different input streams. In this work, we demonstrate a two-block network: the first block is responsible for processing visual data, and the second block is responsible for processing label data. When presented with a complete input (image and label), these two blocks independently learn to predict a 'fused' symbolic representation created by bundling the symbols produced from each input. This 'fused' representation is a symbol which represents both an image and its associated label. Each block also simultaneously learns to recreate its own input from the fused symbol, similar to the process of corollary discharge in the brain [3]. We show that by learning to predict this interaction, the two-block network can be trained to both classify and generate images, and has the potential to be adapted for continual learning.

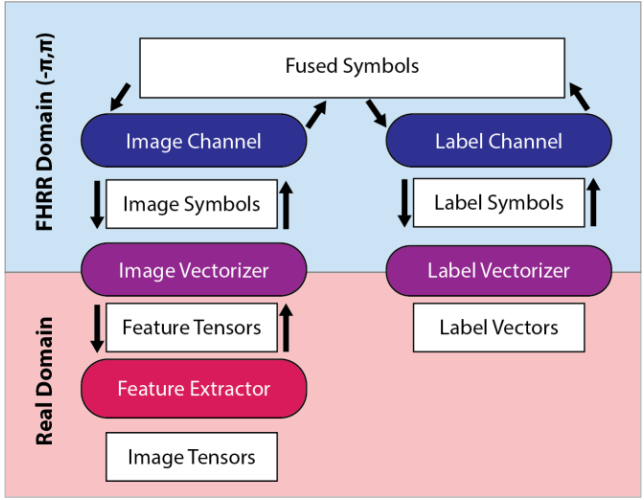

**Figure 1: Illustration of the two-block Bridge network used in this work. In this biologically inspired system, external inputs arrive at the bottom and propagate upwards to be processed. Each 'column' of the system is dedicated to one input stream. The lower layers extract useful features from this input, and the higher layers learn how the features between inputs relate to one another. A complete input consisting of both label and input can be used to train the system, and an incomplete input (image or label) can be used for inference (classification or image generation).**

Each block of a Bridge network is responsible for predicting a 'downstream' or 'output' symbol given an input, and vice-versa, an input given an output symbol. For example, given an image input, the image block extracts its features using a pre-trained convolutional network and randomly projects these data to form a symbol representing this input image. Given this image symbol, the trainable 'channel' at the top of the image block learns to add the information which corresponds to the label symbol it expects to see bundled with this image (Figure 1). And, given an output symbol, it learns to remove the information corresponding to the label to reconstruct only the image symbol. Losses are local for each channel, as each observes only the dissimilarity between its predicted symbols and the true symbols.

In this network, the processes of classification and generation are not trained separately but arise simultaneously from each block learning how its input covaries with the fused symbol. Given only an image, the image block converts it to a symbol as before, and adds to this symbol the information it associates with similar image inputs. This produces an approximation of what the fused symbol would be given the true label at the other block's input. Given this approximate fused symbol, the label block removes the image data it associates with a label to recreate the approximate label symbol, creating information which can be used for classification.

Conversely, given only a label, the opposite route through the network can be taken: the label block adds the image information associated with a label, and the image block removes the associated label information to produce a generated image. We demonstrate these inference modes (classification and generation) in Bridge networks which have been trained on either the standard MNIST or Fashion-MNIST (FMNIST) datasets [13, 27].

## 2.1 Classification

As previously mentioned, a Bridge network performs classification by producing a class symbol which the channels predict is most closely associated with a previously-unseen image input. This predicted symbol can be converted to a 'hard' label by finding its nearest neighbor in the symbol codebook which encodes labels (see methods). This produces a predicted class for each input image without a label. We find that using a pre-trained feature extractor to produce image symbols, the network's classification accuracy on the test set reaches 98.9±0.1% on MNIST and 87.0±0.1% on FMNIST (n=5, dimensionality of vector-symbols $N_D = 1000$). Despite not having classification accuracy directly trained as a loss-related metric in a Bridge network, its performance reaches levels close to the upper limit of a conventional network trained on the same feature extractor (99.1% and 89.5% on MNIST and FMNIST, respectively, Figure 3).

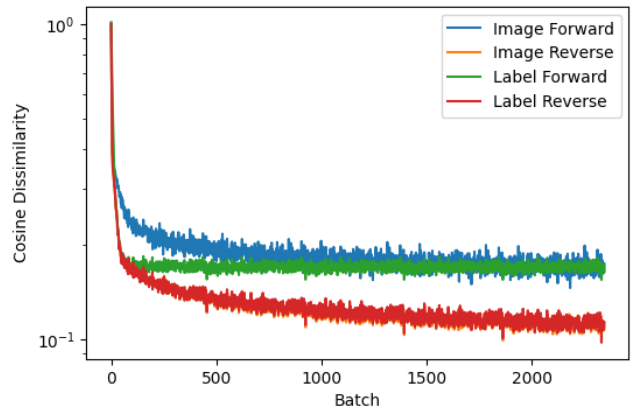

**Figure 2: An example training curve for a Bridge network. The local loss on each channel decreases as it learns to approximate the true fused symbol created by bundling both input symbols given only its own input.**

This classification performance is dependent on the dimension of the vector-symbol used; to reach good performance levels, the symbol must have a dimensionality which allows it to bundle effectively. To show the importance of effective bundling in a Bridge network, we compare it to a control network which does not use a well-defined bundling operator and simply sums symbols on the positive real domain together.

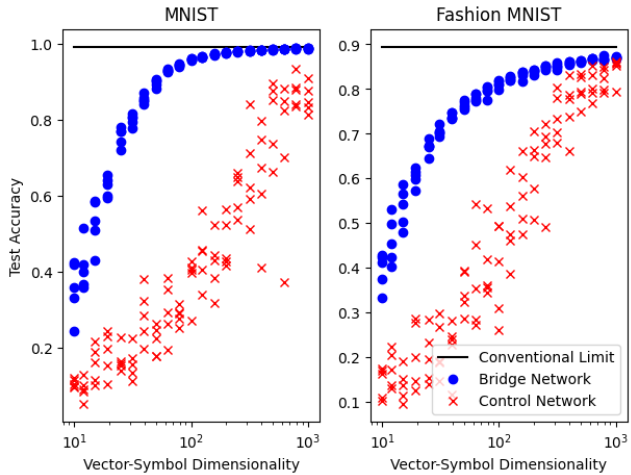

**Figure 3: Bridge network performance on test-set classification after 5 epochs of training. Good performance which reaches the levels of a conventional network (black line) is dependent on using vector-symbols which have sufficiently high dimensionality and a well-defined bundling function (control networks use the same architecture but no VSA operations).**

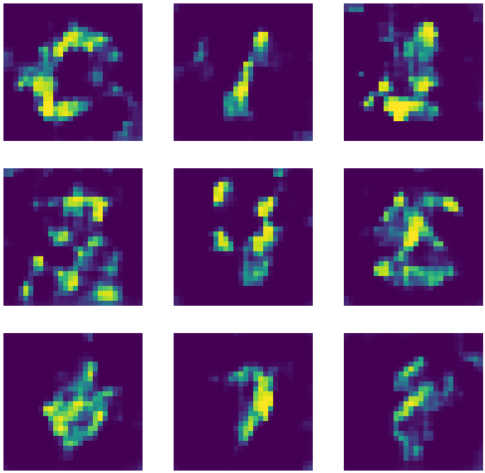

**Figure 4: Examples of digits generated from a Bridge network. Digits contain the major modes of the underlying class but are otherwise currently of low visual quality.**

## 2.2 Generation

In a Bridge network, image generation is another form of inference in which the image symbol is predicted given a label symbol, rather than vice-versa (as used for classification). Backwards operations are also available for the image vectorizer and feature extractor (Figure 1), which allows a full image to be reconstructed. While the obtained images are not of high enough quality to fool a human observer (Figure 4), this ability demonstrates that the network contains information which may be applied to mitigate the problem of catastrophic forgetting via generation replay.

## 2.3 Self-Distillation

Catastrophic forgetting is a phenomenon in which new training data causes disruption to previously-trained tasks in a neural network [12, 16]. One approach to addressing this issue is generative replay, in which new training examples are interleaved with data representing old tasks to prevent performance from being lost on these previous tasks [25]. Although we have shown that images can be generated by the network, generating a suitable set of images on learned classes depends on task-based information which may not be available. To obviate the need for knowledge of the previous and current tasks, we demonstrate the ability of a bridge network to self-distill a set of training examples which can be used in generative replay.

Generative replay may be viewed as a task related to dataset distillation [23, 26]. In both, the task is to produce a set of data which represents a network or model's current 'knowledge.' As the objective of each channel in a bridge network is to add or remove information expected with a certain input, for inputs on which the network has been well-trained the addition and removal of information should approximate inverse operations. That is, by taking a task-relevant image symbol and moving it forward and backward through the image channel, the resulting symbol should be highly similar to the original input. We term the loss of self-similarity through a channel its 'loop loss.'

By selecting a random set of fused symbols which are optimized to simultaneously reduce the loop loss through the image and label channels, a set of fused symbols relevant to currently-learned tasks is obtained. The channels are then used to create the matching image and label symbols for this training set. Using these self-distilled data, a bridge network using the same initial weights can be trained to nearly the same performance level as with the full training dataset.

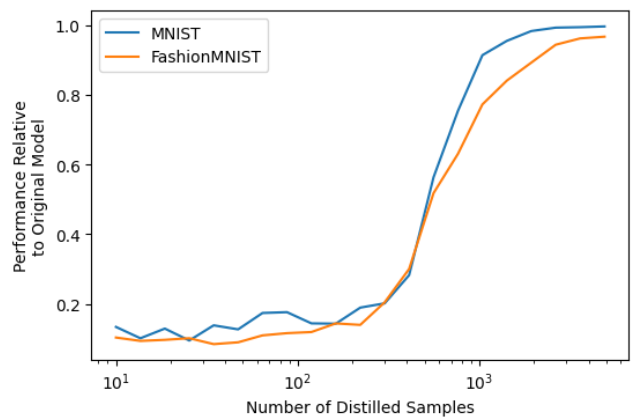

**Figure 5: Classification performance of bridge networks trained from self-distilled data. Performance which is comparable to networks trained with full datasets is achievable, but these networks must share initial weights.**

## 2.4 Continual Learning

We utilize three different scenarios to demonstrate incremental learning of the MNIST image classification task by a bridge network. These three scenarios utilize different sets of information interleaved with new training data to reduce catastrophic forgetting. "Stored Images/Labels" is effectively pseudorehearsal, where all previously seen images and their matching labels are stored and interleaved with new training data.

"Stored symbols" only stores the set of fused symbols resulting from images and labels between tasks, and regenerates these into label and image symbols during the new task. The loss of performance it demonstrates from the previous scenario is likely related to the decrease in similarity between the regenerated and original training symbols due the non-zero loss of the channels' backward operation (reconstructing images/label symbols from fused symbols).

Lastly, "distilled symbols" replaces these fused symbols with ones self-distilled by the network (as described in the last section). This requires only a constant buffer of stored symbols between tasks which is updated after the new training is completed. Final performance is decreased another step by using this method, as the current self-distillation procedure may not provide ideal training points to prevent forgetting. Additionally, in experiments it was observed that with this method interleaving more self-distilled data was required in later tasks to prevent forgetting. However, it still provides much better performance than a control network, which demonstrates complete catastrophic forgetting when sequentially trained on new tasks (Figure 6).

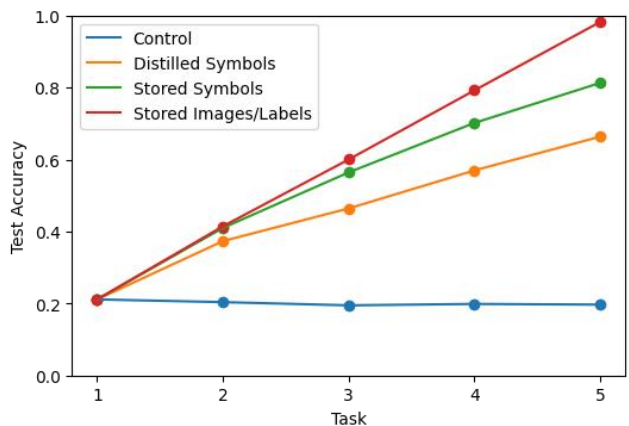

**Figure 6: Performance of a bridge network incrementally learning the digits of the MNIST dataset. Storing all previous inputs addresses catastrophic forgetting through pseudorehearsal (stored images/labels). Storing only the fused symbols from these inputs and regenerating the lower inputs addresses forgetting but final performance is lower (stored symbols). Storing only self-distilled symbols between tasks leads to another drop in final performance but uses constant memory and requires no task data.**

## 3 Discussion

Bridge networks compute using an architecture which differs greatly from conventional deep networks designed for tasks such as image classification and generation. Bridge networks learn the relationships between different data streams represented by vector-symbols, and in doing so can learn to classify and generate images; these two different tasks merely become different aspects of predicting missing information. This shift in architecture allows Bridge networks to learn using local losses and in a biologically-inspired manner, in which information flows from a variety of external sources to be incorporated into internal representations [3].

### 3.1 Continual Learning

Continual learning is an important aspect of biological learning systems which remains a challenge for artificial networks to emulate [12, 25]. We demonstrate that properties of the loss function used in the bridge network can self-distill training examples which can be interleaved with new data to reduce catastrophic forgetting in the network. However, the performance of this technique is not yet comparable with state-of-the-art methods and must be tested on more complex datasets.

### 3.2 Neurosymbolic Computing

The representation of information in the form of a vector-symbol potentially allows Bridge networks to interoperate with a number of complementary vector-symbolic methods and architectures. These methods can provide powerful symbolic reasoning techniques, such as factoring an image into its component pieces [7]. Integrating these methods with Bridge networks could provide a path towards neurosymbolic networks which can both learn from data and symbolically reason.

### 3.3 Multisensory Integration

The results shown in this work currently use a two-channel Bridge network. However, bundling of more than two symbols can be highly effective, and there on this basis there is no reason why more sensory channels cannot be added to provide the integration of a larger number of input streams, such as sound.

## 4 Conclusion

In this work, we have demonstrated the viability of an alternative network architecture. In this architecture, information from external sources is transformed into a symbol defined by a vector-symbolic architecture, the Fourier Holographic Reduced Representation. Symbols produced by different inputs can then be bundled to provide another symbol representing all inputs combined. Each information source in the Bridge network learns to predict this fused symbol given only its own input. This self-predictive act is close to many biologically relevant phenomena and provides an effective basis for executing tasks such as image classification. Furthermore, we believe the unique features of Bridge networks allows them to be extended in future works to address catastrophic forgetting, symbolic reasoning, and more.

# 5 Methods

## 5.1 Feature Extractor

All networks were built in Tensorflow 2.4.1 with Python 3.8.5. The full code and environment required for the results are publicly available at https://github.com/wilkieolin/bridge_networks. The architecture to the convolutional feature extractor and its inverse upsampler is given in Table 1. This network was constructed as an autoencoder and trained to recognize and generate characters from a separate dataset, Kuzushiji MNIST (KMNIST) [4]. Classification was done by adding an additional dense layer (1600x10) at the network's latent space to predict a class label given the extracted features. Classification accuracy reached 92.2% on the KMNIST test set after training for 6 epochs. The weights of the feature extractor and image constructor were then frozen after training and used in inference mode within the MNIST and FashionMNIST bridge networks.

**Table 1: Architecture of convolutional feature extractor and its inverse model.**

| Convolutional Head | | Inverse Convolutional Head | |
|---|---|---|---|
| Input Layer | | Input Layer | |
| Batch Normalization | | Reshape | |
| Conv2D | 32 Filters, 3x3 Kernel | Dropout | 30% |
| MaxPool2D | 2x2 Pool | UpSample2D | 2x2 Pool |
| Dropout | 50% | Conv2DTranspose | 32 Filters, 4x4 Kernel |
| Conv2D | 64 Filters, 3x3 Kernel | Dropout | 50% |
| MaxPool2D | 2x2 Pool | UpSample2D | 2x2 Pool |
| Dropout | 30% | Conv2DTranspose | 1 Filter, 3x3 Kernel |
| Flatten | | | |

## 5.2 Image & Label Vectorizers

Tensors of image features calculated from the pre-trained feature extractor exist on the positive real domain, and were converted to the domain used by the FHRR VSA by random projection, a $3\sigma$ batch normalization, and clipping. This transforms the inputs on the domain $(0, \infty)$ to have 99% of inputs fall within the domain [-1, 1] and the remainder of extreme values clipped within this domain. The reverse operation is accomplished by inverting the batch-normalization process and multiplying the resulting output by the random projection's pseudoinverse.

Vectors of integral image labels were converted to FHRR VSA symbols via codebook: each class is assigned a random hypervector symbol to represent it in the FHRR domain. Small amounts of noise can be added to this symbol to produce variability in generated images.

## 5.3 Channels

Channels are identical for different inputs and consist of two networks performing a forward and reverse transform. The forward and reverse networks are each a single residual block, consisting of a hidden dense layer and an output layer which adds the hidden layer's output and the equally-sized input. The residual block is used here as it can easily approximate the identity function: if there is no mutual information between the input channels, a channel will ideally produce the identity function. Otherwise, the forward channel will learn to add the information which it expects to see from the other channel, and the reverse channel will learn to remove it. During training of the 2-channel Bridge networks demonstrated above, these channels are the only portions of the network being trained via backpropagation; all other blocks are frozen (feature extractors) or static/internally training (image vectorizer).

## ACKNOWLEDGMENTS

Acknowledgement. This work was supported by NIH T-32 Training Grant (5T32MH020002), the Lifelong Learning Machines program from DARPA/MTO (HR0011-18-2-0021) and ONR (MURI: N00014-16-1-2829).